# Digital Twins for forecasting and decision optimisation with machine learning: applications in wastewater treatment

Matthew Colwell [1], Mahdi Abolghasemi [2]

School of Mathematics and Physics, The University of Queensland

Email: [1] matthew.colwell@uq.edu.au, [2] m.abolghasemi@uq.edu.au

**Introduction**

Prediction and optimisation are two widely used techniques that have found many applications in solving real-world problems. While prediction is concerned with estimating the unknown future values of a variable, optimisation is concerned with optimising the decision given all the available data. These methods are used together to solve problems for sequential decision-making where often we need to predict the future values of variables and then use them for determining the optimal decisions. This paradigm is known as "forecast and optimise" and has numerous applications, e.g., forecast demand for a product and then optimise inventory, forecast energy demand and schedule generations, forecast demand for a service and schedule staff, to name a few. In this extended abstract, we review a digital twin that was developed and applied in wastewater treatment in Urban Utility to improve their operational efficiency. While the current study is tailored to the case study problem, the underlying principles can be used to solve similar problems in other domains.

**Purpose**

This paper presents a novel methodology applied to a sequential decision-making problem with competing priorities and differing levels of importance.

**Case study**

Urban Utilities (UU) provides wastewater treatment services to the south-east Queensland region. Over the past year, we have been working with Urban Utilities' (UU) Resource Recovery team whose responsibilities concern the safe and efficient operation of the company's wastewater treatment plant (WWTP) infrastructure. We have implemented a project with this team help operate their wastewater treatment equipment at lower cost, by applying a data-driven approach.

The wastewater treatment process generates solid waste, typically referred to as biosolids [4]. At Urban Utilities, biosolids are stabilised by specialised reactors at their Oxley WWTP. This equipment is colloquially referred to by its manufacturer's name, Cambi. The End of Waste Code (part of the *Waste Reduction and Recycling Act, 2011*) requires that biosolids generated from wastewater treatment, which are a biological hazard, are treated prior to reuse or disposal [9]. Cambi achieves the necessary treatment of biosolids by thermal hydrolysis (destruction by reaction with water) [5]. Thermal hydrolysis sterilises the biosolids and produces a stable product which is suitable for agricultural reuse. However, thermal hydrolysis involves a lot of energy over a brief period (up to an hour), after which the biosolids are typically left in digestion for several weeks [7].

Cambi is currently operated based on heuristics (and the intuition of the plant operators at Oxley) and therein laid the opportunity for this project. The Cambi equipment is complex and subject to several logistical constraints: operation of the system is viewed more as an art than a science, with varying opinions about how to run it. Running Cambi is one of UU's single largest operational costs, so small efficiency gains could result in a significant cost saving [8]. Our primary objective in this project was to help realise this cost saving by using a data-driven forecast and optimise approach to operation.

There are multiple objectives for the Cambi system, however, under the current intuition-based operation, operators are forced to prioritise some objectives and disregard others, due to the system's complexity. In order of priority, the objectives within the Cambi system are, based on communication with UU employees:

- to maintain a low level in the upstream storage such that it can accept new trucked deliveries, via appropriate throughput management;
- to produce an acceptable biosolid quality, within the specification of the End of Waste Code (this is a lower priority than flow since this is rarely an issue); and finally,
- to produce at the lowest possible cost.

With rising gas prices [5], this paper highlights the need to further investigate this final objective and provides a framework to do so, using the data available to UU.

**Data**

The data, which is available to UU includes historical plant data, accessible through UU's various data warehouses [10], and Bureau of Meteorology weather data. Exploration of the data revealed several correlations, which could be exploited for efficiency gains, and regression models demonstrated the non-linearity of the data.

**Methodology**

To build a decision support system which preserves the current operating objective philosophy of Cambi, we first constructed a mixed-integer programming (MIP) model whose objective function minimises the difference between the level in the upstream storage unit and a target level, over some planning horizon. This ensures that throughput is still the primary objective of the recommendation system. Next, we applied a novel machine learning regression model, which predicts the behaviour of the set of all possible operating scenarios, subject to the constraints described by the MIP outputs. This model reports the case within this set which it predicts is the most efficient, contrary to the conventional predict and optimise framework. This structure allowed us to use the regression model to predict both biosolid quality and energy efficiency, choosing the inputs which resulted in the lowest energy usage, while still subject to the constraint of acceptable biosolid quality. The MIP optimisation model and the machine learning regression model together form the basis for the decision optimisation system, which collates live operation data and provides feedback to the operators.

The MIP model was implemented using Google's Python based OR-Tools library and minimises the function $\sum_{t=0}^{T} z_t + \omega \sum_{t=0}^{T-1} \sum_{r=1}^{3} y_{r,t}$ where $z_t$ is the absolute difference between the level in the upstream storage unit and a target level, $\omega$ is a weighting factor determined by trial-and-error, and $y_{r,t}$ is a binary variable representing if reactor $r$ was turned on/off in time step $t$. The first term of the objective function ensures that the level never deviates too far from the target and the second term ensures that operator intervention is not laborious. Figure 1 shows the performance of the MIP model in comparison to the current manual operation and it can be seen the MIP model maintains a much more stable storage level.

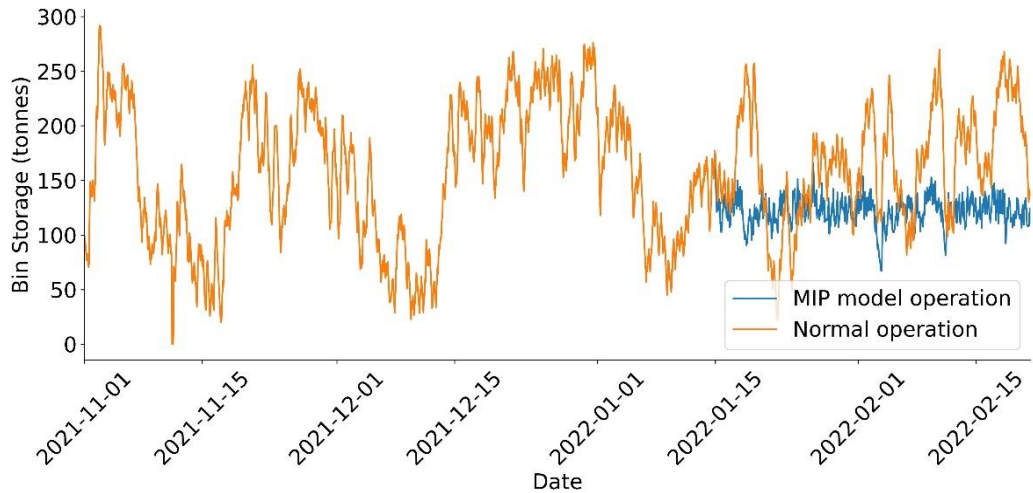

*Figure 1: MIP Model Performance Against Actual Operation Level*

The machine learning regression model was developed by testing a shortlist of candidate architectures including SVM (with RBF kernel), AdaBoost, Random Forest, LightGBM, kNN and MLP. While not a native multi-output model, LightGBM, with a multi-output wrapper from the Scikit-learn library, was found to predict Cambi's behaviour on a test set better than the other models [3,6].

It is estimated that if UU were to adopt the recommendations of this project, they could save tens of thousands of dollars significantly on natural gas consumption. We implemented the forecasting and optimisation models in Python and developed a dashboard in PowerBI for visualisation.

**Limitations**

In this current study we developed a forecasting and an optimisation model for the problem and separately applied them to provide a solution. However, these two phases can be integrated into a single model where the forecasting model considers the quality of the final decisions [1,2]. Thus, the forecasting model generates more accurate forecasts which will be used for improving the quality of decisions in the optimisation model. The whole process can be also integrated into an end-to-end model to directly map the inputs to final optimal decisions.

**Conclusion**

We conclude this paper by reiterating the cost savings that a data driven approach with machine learning to wastewater treatment plant operation can provide. Furthermore, we highlight where the proposed "forecast and optimise with machine learning" approach could be extended beyond Cambi. Finally, we encourage further research in designing of digital twins for forecasting and decision optimisation with machine learning models, especially in automating the process and further improving the quality of forecasts and decisions by integrating them.

# Reference

[1] *Abolghasemi, M., Abbasi, B., Babaei, T., & HosseiniFard, Z. (2021). How to effectively use machine learning models to predict the solutions for optimization problems: lessons from loss function. arXiv preprint arXiv:2105.06618..*

[2] Abolghasemi, M., & Esmaeilbeigi, R. (2021). *State-of-the-art predictive and prescriptive analytics for IEEE CIS 3rd Technical Challenge.* arXiv preprint arXiv:2112.03595.

[3] *Abolghasemi, M., Tarr, G., & Bergmeir, C. (2022). Machine learning applications in hierarchical time series forecasting: Investigating the impact of promotions. International Journal of Forecasting.*

[4] Australian Government. (2022). *Gas Market Prices*. Australian Energy Regulator.

https://www.aer.gov.au/wholesale-markets/wholesale-statistics/gas-market-prices

[5] Burton, F.L., Stensel, H.D., Tchobanoglous, G., Tsuchihashi, R. (2013). *Wastewater Engineering: Treatment and Resource Recovery 5th ed*. New York: McGraw-Hill

[6] Cambi. (2021). How does thermal hydrolysis work? Retrieved from https://www.cambi.com/what-we-do/thermal-hydrolysis/how-does-thermal-hydrolysis-work/

[7] Dwyer, J., Starrenburg, D., Tait, S., Barr, S., Batstone, D., Lant, P. (2008). Decreasing activated sludge thermal hydrolysis temperature reduces product colour, without decreasing degradability. *Water Research*, *42*, 4699-4709.

[8] Malekizadeh, A. (2019, May). *Thermal Hydrolysis Process Optimisation in Relation to Anaerobic Digestion*. Paper presented at OzWater 2019 in Melbourne, Australia.

[9] *Waste Reduction and Recycling Act 2011* (Qld). Retrieved from https://www.legislation.qld.gov.au/view/pdf/inforce/current/act-2011-031

[10] Wong, R. (2020). *ProcessDataLink Management Procedure*. Urban Utilities.

*Acknowledgment:*

*In addition to the above, personal communication with employees of Urban Utilities has been instrumental in developing the problem definition and discussions with Dr. Mahdi Abolghasemi have been vital to developing modelling ideas for this problem.*